\documentclass[a4paper,twoside]{article}

\usepackage{epsfig}
\usepackage{subfigure}
\usepackage{calc}
\usepackage{amssymb}
\usepackage{amstext}
\usepackage{amsmath}
\usepackage{amsthm}
\usepackage{multicol}
\usepackage{pslatex}
\usepackage{apalike}
\usepackage{SCITEPRESS}     

\begin{document}

\title{Automatic Information Extraction from Piping and Instrumentation Diagrams}

\author{\authorname{Rohit Rahul, Shubham Paliwal, Monika Sharma and Lovekesh Vig}
\affiliation{TCS Research, New Delhi, India}
\email{\{rohit.rahul, shubham.p3, monika.sharma1, lovekesh.vig\}@tcs.com}
}


\abstract{One of the most common modes of representing engineering schematics are Piping and Instrumentation diagrams (P\&IDs) that describe the layout of an engineering process flow along with the interconnected process equipment. Over the years, P\&ID diagrams have been manually generated, scanned and stored as image files. These files need to be digitized for purposes of inventory management and updation, and easy reference to different components of the schematics. There are several challenging vision problems associated with digitizing real world P\&ID diagrams.  Real world P\&IDs come in several different resolutions, and often contain noisy textual information. Extraction of instrumentation information from these diagrams involves accurate detection of symbols that frequently have minute visual differences between them. Identification of pipelines that may converge and diverge at different points in the image is a further cause for concern. Due to these reasons, to the best of our knowledge, no system has been proposed for end-to-end data extraction from P\&ID diagrams. However, with the advent of deep learning and the spectacular successes it has achieved in vision, we hypothesized that it is now possible to re-examine this problem armed with the latest deep learning models. To that end, we present a novel pipeline for information extraction from P\&ID sheets via a combination of traditional vision techniques and state-of-the-art deep learning models to identify and isolate pipeline codes, pipelines, inlets and outlets, and for detecting symbols. This is followed by association of the detected components with the appropriate pipeline.  The extracted pipeline information is used to populate a tree-like data structure for capturing the structure of the piping schematics. We have also evaluated our proposed method on a real world dataset of P\&ID sheets obtained from an oil firm and have obtained extremely promising results. To the best of our knowledge, this is the first system that performs end-to-end data extraction from P\&ID diagrams.}

\onecolumn \maketitle \normalsize \vfill

\begin{figure*}[t]
\begin{center}
   \includegraphics[width=1\linewidth]{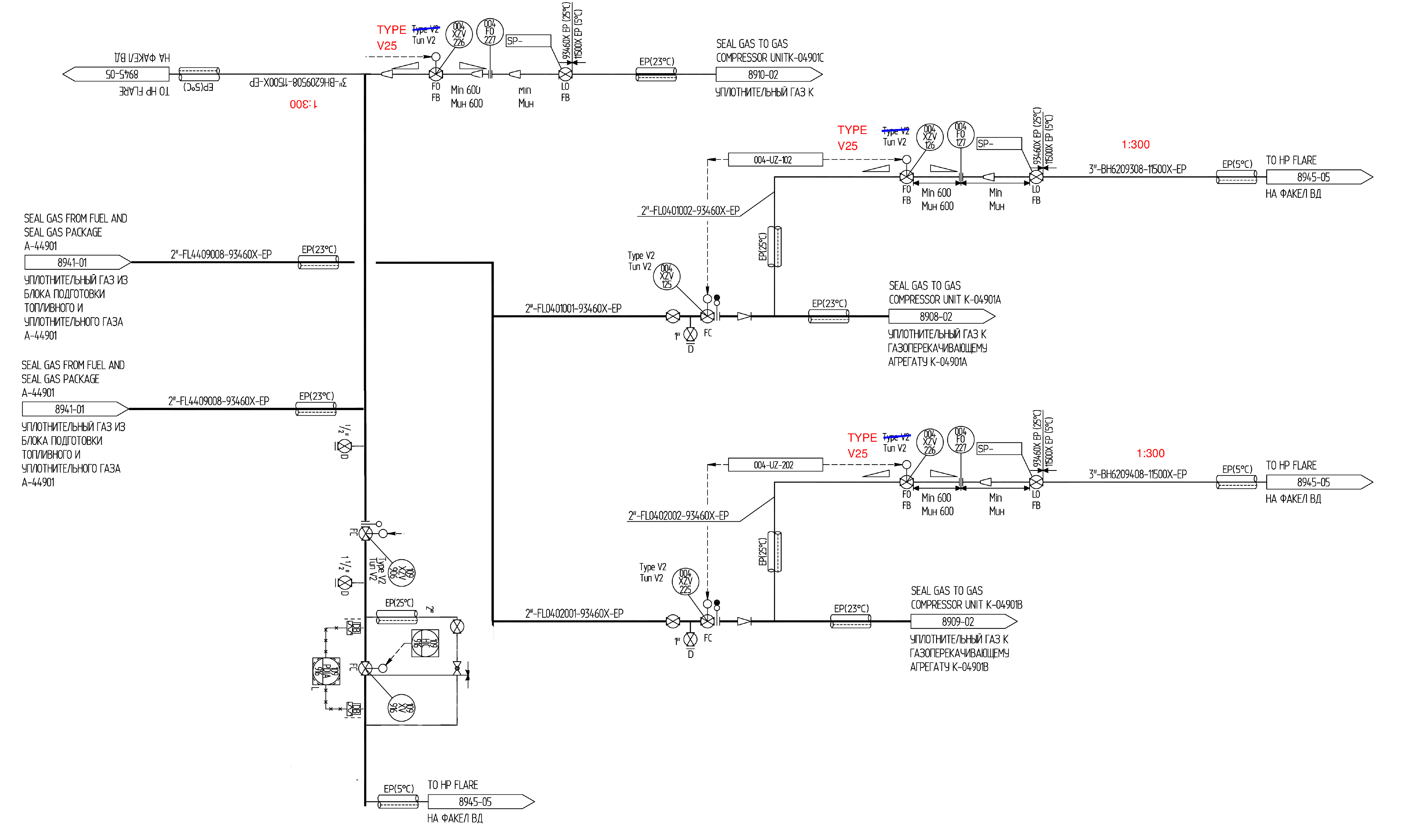}
\end{center}
   \caption{An Example of Piping and Instrumentation Diagram sheet.}
\label{fig:sample-sheet}
\end{figure*}

\vspace{-2mm}
\section{Introduction}
\label{sec:intro}
\vspace{-2mm}
A standardized representation for depicting the equipment and process flow involved in a physical process is via Piping and Instrumentation diagrams (P\&ID). P\&ID diagrams are able to represent complex engineering workflows depicting schematics of a process flow through pipelines, vessels, actuators and control valves. A generic representation includes fluid input points, paths as pipelines, symbols which represent control and measurement instruments and, sink points. Most industries maintain these complex P\&IDs in the form of hard-copies or scanned images and do not have any automated mechanism for information extraction and analysis of P\&IDs~\cite{pid_2}. Consequently, future analysis and audit for process improvement involves manual involvement which is expensive given the domain expertise required. It would be of great value if the data present in P\&ID sheets could be automatically extracted and provide answers to important queries related to the connectivity of plant components, types of interconnections between process equipments and the existence of redundant paths automatically. This would enable process experts to obtain the information instantly and reduce the time required for data retrieval. Given the variations in resolution, text fonts, low inter-class variation and the inherent noise in these documents, this problem has previously been considered too difficult to address with standard vision techniques. However, deep learning has recently shown incredible results in several key vision tasks like segmentation, classification and generation of images. The aim of this paper is to leverage the latest work in deep learning to address this very challenging problem, and hopefully improve the state-of-the-art for information extraction from these P\&ID diagrams.  

The digitization process of P\&IDs involves identification and localization of pipeline codes, pipelines, inlets, outlets and symbols which is followed by mapping of individual components with the pipelines. Although tools for the digitization of engineering drawings in industries are in high demand, this problem has received relatively little attention in the research community. Relatively few attempts have been made in the past to address digitization of complex engineering documents comprising of both textual and graphical elements, for example: complex receipts, inspection sheets, and engineering diagrams~\cite{verma2016automatic},~\cite{arrow_markings},~\cite{pid_2},~\cite{mhps_cbdar},~\cite{symbol}. We have found that connected component analysis~\cite{koo_cc} is heavily employed for text-segmentation for such documents~\cite{verma2016automatic}. However, the recently invented Connectionist Text Proposal Networks (CTPN)~\cite{ctpn} have demonstrated the capability to detect text in extremely noisy scenarios. We utilize a pre-trained CTPN network to accurately detect the text patches in a 
P\&ID image.  In the literature, symbol detection is performed by using shape based matching techniques~\cite{shape_based_matching}, auto associative neural networks~\cite{symbol_1}, graph based techniques~\cite{symbol_2}. However, detecting symbols in P\&ID sheets is quite challenging because of the low inter-class variation among different symbols and the presence of text and numbers inside symbols. To alleviate this issue, we succesfully employ Fully Convolutional Networks (FCN) which are trained to segment out the individual symbols.

Thus, our proposed pipeline for information extraction from P\&ID sheets uses a combination of state-of-the-art deep learning models for text and symbol identification, in combination with low level image processing techniques for the  extraction of different components like inlets, outlets and pipelines present in the sheets. Moreover, given the paucity of sufficient real datasets for this domain, automating the process of information extraction from P\&ID sheets is often harder than in other domains and significant data augmentation is required to train deep models. We evaluate the efficacy of our proposed method on 4 sheets of P\&IDs, each containing multiple flow diagrams, as shown in Figure~\ref{fig:sample-sheet}. 

To summarize, we have formulated the digitization process of P\&IDs as a combination of (1) heuristic rule based methods for accurate identification of pipelines, and for determining the complete flow structure and (2) deep learning based models for identification of text and symbols and (3) rule based association of detected objects and a tree based representation of process flow followed by pruning for determining correct inlet to outlet path. While formulating the digitization process of P\&IDs, we make the following contributions in this paper:

\begin{figure*}[t]
\begin{center}
   \includegraphics[width=1\linewidth]{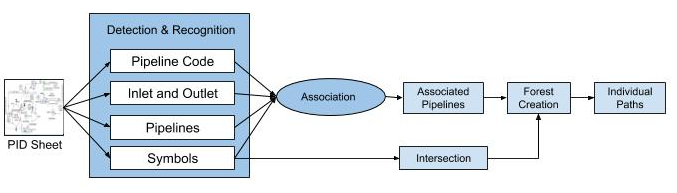}
\end{center}
   \caption{Flowchart showing proposed 2-step process for digitization of Piping and Instrumentation Diagrams. First, P\&ID sheet is fed to a detection and recognition engine which identifies and isolates different components of the process flow like pipelines, pipeline codes, inlets, outlets and symbols using a combination of traditional vision techniues and deep learning models. Subsequently, the extracted components are sent to an association module for mapping with the appropriate pipeline. Finally, a tree-like data structure is created to determine the flow from inlet to outlet.}
\label{fig:flowchart}
\end{figure*}

\begin{itemize}
\item We propose a novel pipeline consisting of a two-step process for information extraction from P\&ID diagrams, comprising of a combination of detection of different components of the process flow followed by their association with appropriate pipeline and representation in a tree-like data structure to determine the flow from inlet to outlet.
\item We propose the use of conventional image processing and vision techniques to detect and recognize graphic objects (e.g. pipelines, inlets and outlets) present in P\&ID.
\item We use a fully convolutional neural network (FCN) based segmentation for detection of symbols in P\&ID sheets at the pixel level because of very minute visual difference in appearance of different symbols, as the presence of noisy and textual information inside symbols makes it difficult to classify based on bounding box detection networks like Faster-RCNN~\cite{faster-rcnn}.
\item We evaluate our proposed pipeline on a dataset of real P\&ID sheets from an oil firm and present our results in Section~\ref{sec:experiment-results}.
\end{itemize}

The remainder of the paper is organized as follows: Section~\ref{sec:related-work} gives an overview of related work in the field of information extraction from visual documents. An overview of the proposed pipeline for automatic extraction of information from P\&ID is given in Section~\ref{sec:overview}. Section~\ref{sec:proposed-methodology} describes in detail the proposed methodology for extracting different P\&ID components like pipeline code, pipelines and symbols etc. and their mapping. Subsequently, Section~\ref{sec:experiment-results} gives details about the dataset, experiments and a discussion on the obtained results. Finally, we conclude the paper in Section~\ref{sec:conclusion}.

\section{Related Work}
\label{sec:related-work}
There exists very limited work on digitizing the content of engineering diagrams to facilitate fast and efficient extraction of information. The authors \cite{engineering_drawing} automated the assessment of AutoCAD Drawing Exchange Format (DXF) by converting DXF file into SVG format and developing a marking algorithm of the generated SVG files. A framework for engineering drawings recognition using a case-based approach is proposed by ~\cite{case-based-ed} where the user interactively provides an example of one type of graphic object in an engineering drawing and then system tries to learn the graphical knowledge of this type of graphic object from the example and later use this learned knowledge to recognize or search for similar graphic objects in engineering drawings. Authors of ~\cite{pid_1} tried to automate the extraction of structural and connectivity information from vector-graphics-coded engineering documents. A spatial relation graph (SRG) and its partial matching method are proposed for online composite graphics representation and recognition in ~\cite{pid_3}. Overall, we observed that there does not exist much work on information extraction from plant engineering diagrams. 

However, we discovered a significant body of work on recognition of symbols in prior art. ~\cite{symbol} proposed Fourier Mellin Transform features to classify multi-oriented and multi-scaled patterns in engineering diagrams. Other models utilized for symbol recognition include Auto Associative neural networks ~\cite{symbol_1}, Deep Belief networks~\cite{symbol_3}, and consistent attributed graphs (CAG)~\cite{symbol_2}. There are also models that use a set of visual features which capture online stroke properties like orientation and endpoint location~\cite{symbol_4}, and shape based matching between different symbols~\cite{shape_based_matching}. We see that most of the prior work focuses on extracting symbols from such engineering diagrams or flow charts. To the best of our knowledge, there exists no work which has proposed an end-to-end pipeline for automating the information extraction from plant engineering diagrams such as P\&ID.

In literature, Connected Component (CC) analysis~\cite{koo_cc} has been used extensively for extracting characters~\cite{mhps_cbdar} from images. However, connected components are extremely sensitive to noise and thresholding may not be suitable for P\&ID text extraction. Hence, we utilize the recently invented Connectionist Temporal Proposal Network (CTPN)~\cite{ctpn} to detect text in the image with impressive accuracy. For line detection, we utilize Probabilistic hough transform (PHT)~\cite{hough_transform} which is computationally efficient and fast version of the standard hough transform as it uses random sampling of edge points to find lines present in the image. We make use of PHT for determining all the lines present in P\&ID sheets which are possible candidates for pipelines. In our paper, we propose the use of Fully convolutional neural network (FCN) based segmentation~\cite{shelhamerLD16} for detecting symbols because tranditional classification networks were unable to differentiate among different types of symbols due to very minute inter-class differences in visual appearances and presence of noisy and textual information present inside symbols. FCN incorporates contextual as well as spatial relationship of symbols in the image, which is often necessary for accurate detection and classification of P\&ID symbols. 

\section{Overview}
\label{sec:overview}

The main objective of the paper is to extract the information from the P\&ID sheets representing schematic process flow through various components like pipelines, valves, actuators etc. The information is extracted from P\&ID and stored in a data structure that can be used for querying. The P\&ID diagram shown in Figure~\ref{fig:sample-sheet} depicts the flow of oil through pipelines from inlet to outlet, where outlets and inlets denote the point of entry and exit of the oil, respectively. Each outlet is unique and may connect to multiple inlets, forming a one-to-many relationship. The symbols indicate the machine parts present on the pipeline to control the flow and to filter the oil in a specific way. The pipelines are identified by a unique P\&ID code which is written on top of every pipeline.

To capture all the information from the P\&ID sheets, we propose a two-step process as follows :
\begin{itemize}
\item In the first step, we identify all the individual components like pipelines, pipeline codes, symbols, inlets and outlets. We use conventional image processing and vision techniques like connected component analysis~\cite{koo_cc}, probabilistic hough transform~\cite{hough_transform}, geometrical properties of components etc. to localize and isolate pipelines, pipeline codes, inlets and outlets. Symbol detection is carried out by using fully convolutional neural network based segmentation~\cite{shelhamerLD16} as symbols have very minute inter class variations in visual appearances. Text detection is performed via a Connectionist Text Proposal Network (CTPN), and the recognition is performed via the tesseract OCR library.   
\item In the second step, we associate these components with each other and finally capture the flow of oil through pipelines by forming a tree-like data structure. The tree is able to represent one-to-many relationship where each outlet acts as root node and each inlet is treated as a leaf node. The pipelines represent intermediate nodes present in the tree.

\end{itemize}

\section{Proposed Methodology}
\label{sec:proposed-methodology}
In this section, we discuss the proposed methodolody for extracting information from P\&ID sheets in detail. It is a two-step process as shown in Figure~\ref{fig:flowchart} in which the first step involves detection and recognition of individual components like pipeline-codes, symbols, pipelines, inletss and outlet. The second step involves association of detected components with the appropriate pipelines followed by formulation of tree-like data structure for finding the process flow of pipeline schematics. These steps are detailed as follows :

\subsection{Detection and Recognition}
\label{subsec:detection}
We use vision techniques for extracting different components like pipeline-codes, symbols, pipelines, inlets and outlets present in P\&IDs. We divide these components into two-types : 1. text containing pipeline-codes and 2. graphic objects like pipelines, symbols. As observed from Figure~\ref{fig:sample-sheet}, P\&ID sheets have text present which represents pipeline code, side notes, sometimes as part of a symbol or container / symbol / tag numbers, we call these text segments as \textit{pipeline-code}. The non-text components like pipelines, symbols, inlets and outlets are termed as \textit{graphic objects}. Now, we discuss the detection and recognition methods for different components as follows :
\begin{itemize}
\item \textbf{Detection of Pipeline code} : The pipeline code distinctly characterizes each pipeline. Hence, we first identify the pipeline code. While previous approaches utilized thresholding followed by connected components in order to extract the codes, we utilize a CTPN~\cite{ctpn} network pre-trained on a scene-text dataset for pipeline-code detection, as it was far more robust to noise / color in the document. CTPN is a convolutional network which accepts arbitrarily sized images and detects a text line in an image by densely sliding a window in the convolutional feature maps and produces a sequence of text proposals. This sequence is then passed through a recurrent neural network which allows the detector to explore meaningful context information of text line and hence, makes it powerful to detect extremely challenging text reliably. The CTPN gives us all possible candidate components for pipeline code with 100 \% recall but with significant number of false positives which are filtered out in a later step. Subsequently, we use tesseract~\cite{tesseract} for reading each component detected in the previous step. Since, pipeline codes have fixed length and structure, we filter out false positives using regular expressions. For example, the pipeline code is of the format N"-AANNNNNNN-NNNNNA-AA where N denotes a Digit and A denotes an alphabet. This domain knowledge gives us all the pipeline codes present in the P\&ID sheets.\\

\item \textbf{Detection of Inlet and Outlet} : The inlet or outlet marks the starting or ending point of the pipeline. There is a standard symbol representing inlet or outlet. It is a polygon having 5 vertices and the width of the bounding box is at least thrice its height. We use this shape property of the symbol to detect inlet / outlet robustly using heuristics. For detection of the inlets and outlets, we subtract the text blobs detected as pipeline codes from a binarized input image for further processing. Then, we use Ramer-Douglas algorithm~\cite{ramer-d} in combination with known relative edge lengths to identify the polygons. After detecting each polygon, we find out whether it is an inlet or an outlet. As can be seen from Figure~\ref{fig:sample-sheet}, there are 4 possible cases of polygons because there are two types of tags present in P\&ID : left-pointing and right-pointing. Each of the right-pointing or left-pointing tag can either be an inlet or an outlet. We find the orientation of tags from the points given by Ramer-Douglas knowing the fact that there will be 3 points on one side and two on another side in a right-pointing or left-pointing tag, as shown in Figure~\ref{fig:tags}. To further classify whether the candidate is an inlet or outlet among them, we take a small kernel K on either side of the component image and find out which edge is crossed by a single line.\\

\begin{figure}[h]
\begin{center}
   \includegraphics[width=0.9\linewidth]{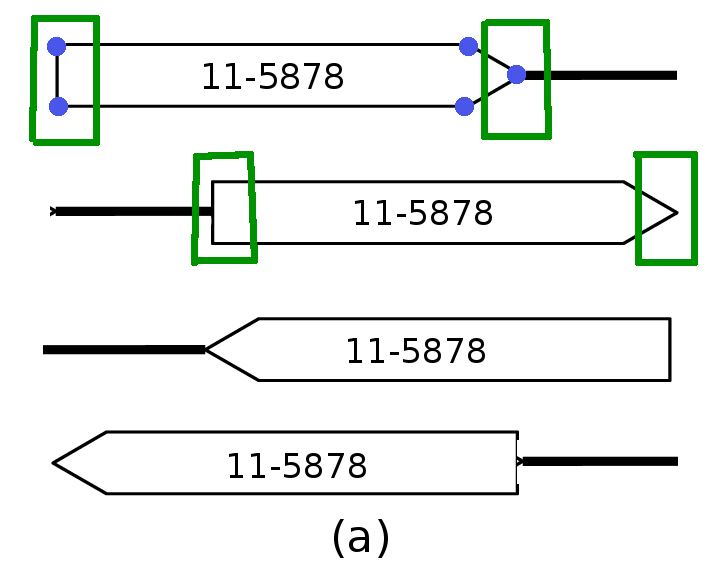}
\end{center}
   \caption{Figure showing inlets and outlets of P\&ID diagrams.}
\label{fig:tags}
\end{figure}

\begin{figure}[t]
\begin{center}
   \includegraphics[width=0.9\linewidth]{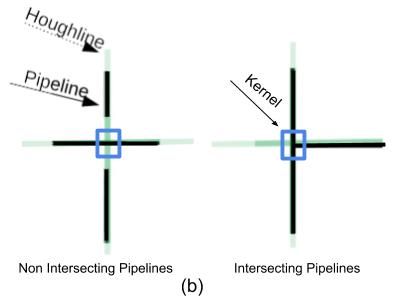}
\end{center}
   \caption{Figure showing pipelines in P\&ID sheets}
\label{fig:intersecting-lines}
\end{figure}

\item \textbf{Detection of Pipeline} : We remove the detected text and inlet / outlet tags from the image for detecting pipelines. We then use probabilistic hough transform~\cite{hough_transform} on the skeleton~\cite{skeleton} version of the image which outputs a set of all lines including lines that do not correspond to pipelines.\\
 
\item \textbf{Detection of Pipeline Intersections} : The output of the hough lines is a set of lines which does not take into account the gap at the intersections, as shown in Figure~\ref{fig:intersecting-lines}. There can be two kinds of intersections : a valid intersection or an invalid intersection. We aim to find all the valid intersections. This is achieved by determining all the intersections between any two line segments by solving the system of linear equations. The solution to the equations is a point which should lie on both finite pipelines. This assumption ensures that the solution is a part of foreground. 
An invalid intersection is the intersection where the solution of the two linear equations for the line has given us an intersection but there exists no such intersection in the image. This is indicated by the gap in one of the lines involved in the intersection, as shown in Figure~\ref{fig:intersecting-lines}. To discard invalid intersections, we draw a square kernel of size 21 with the center at the intersection and check for lines intersecting with the edges of the square. Here, we have two possibilities : (1) where the intersections are on the opposite edges of the square  and no intersection on other two edges of the square. This means that there is no intersection and there is just one line which passes through the intersection. (2) where there can be intersection on three or all four edges of the square. This is the case of valid intersection between the pipelines. Thus we obtain the pipeline intersections and store them for use later to create of a tree-like data structure for capturing the structure of pipeline schematics.\\

\item \textbf{Detection of Symbols} : There are various types of symbols present in the P\&ID sheets which represent certain instruments responsible for controling the flow of oil through pipelines and performing various tasks. In our case, we have 10 classes of symbols to detect and localise in the sheets, e.g. ball\_valve, check\_valve, chemical\_seal, circle\_valve, concentric, flood\_connection, globe\_valve, gate\_valve\_nc, insulation and globe\_valve\_nc. As can be seen in Figure~\ref{fig:symbols}, these symbols have very low inter-class difference in visual appearances. So, standard deep networks for classification are not able to distinguish them correctly. Therefore, we propose to use fully convolutional neural network (FCN)~\cite{shelhamerLD16} for detecting symbols. FCNs, as shown in Figure~\ref{fig:fcn}, are convolutional networks where the last fully connected layer is replaced by a convolution layer with large receptive field. The intuition behind using segmentation is that FCN network has two parts : one is downsampling path which is composed of convolutions, max pooling operations and extracts the contextual information from the image, second is upsampling path which consists of transposed convolutions, unpooling operations to produce the output with size similar to input image size and learns the precise spatial location of the objects in the image. \\

\begin{figure}[t]
\begin{center}
   \includegraphics[width=0.9\linewidth]{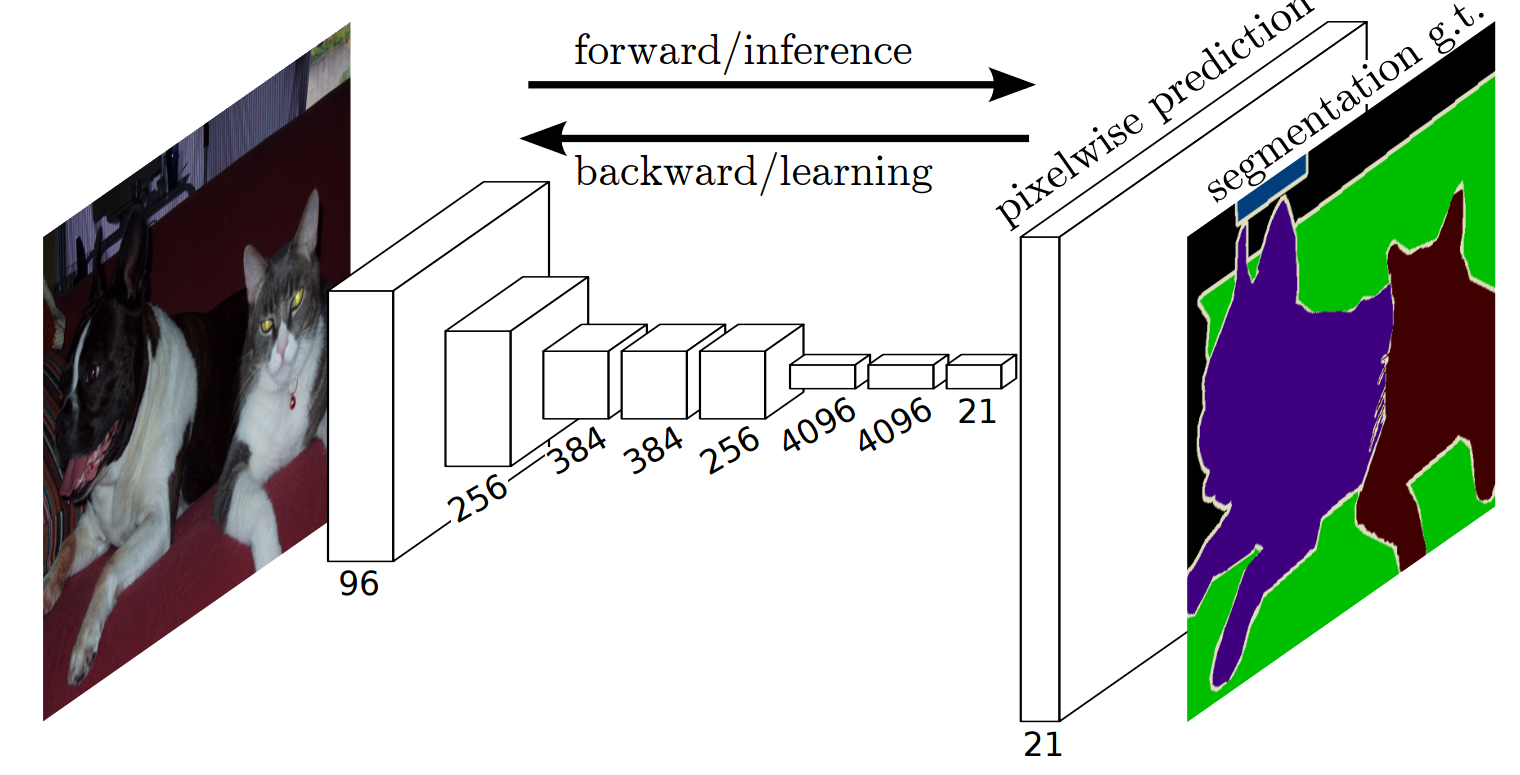}
\end{center}
   \caption{Fully convolutional segmentation network taken from ~\cite{shelhamerLD16}}
\label{fig:fcn}
\end{figure}

\begin{figure*}[t]
\begin{center}
   \includegraphics[width=0.8\linewidth]{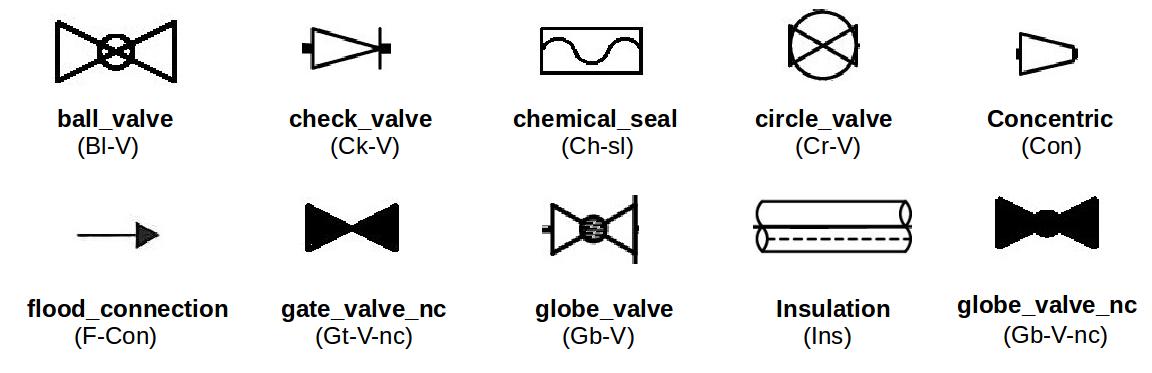}
\end{center}
   \caption{Different classes of symbols present in P\&ID sheets.}
\label{fig:symbols}
\end{figure*} 

\noindent \textbf{Data Annotation for FCN} : For detecting symbols using FCN, we annotated a dataset of real world P\&IDs diagrams from an oil firm. The original P\&ID sheets are of very large size, so we divided it into smaller patches of size $400 \times 400$ for annotating the symbols. These patches contain different classes of symbols and can have multiple symbols present in a single patch. The symbols were annotated by masking their pixel values completely and subsequently, obtaining the boundaries of the symbol masks representing the shape of the symbol. To automate this process of extracting outlines of symbol masks, a filter was applied for the region containing the masked shape, i.e, bitwise-and operation was used. This was followed by thresholding the patches to get the boundaries / outlines only and then it was dilated with a filter of size $3 \times 3$. As the training dataset was limited, we augmented the images by performing some transformations on the image like translation and rotation.\\

\noindent \textbf{Training Details} : We use VGG-19~\cite{vgg19} based FCN for training symbol detector. An input image of size $400 \times 400$ is fed to the network and it is trained using Adam optimizer with a learning rate of 0.0004 and batch size of 8.

%

\end{itemize}

\begin{figure}[t]
\begin{center}
   \includegraphics[width=0.9\linewidth]{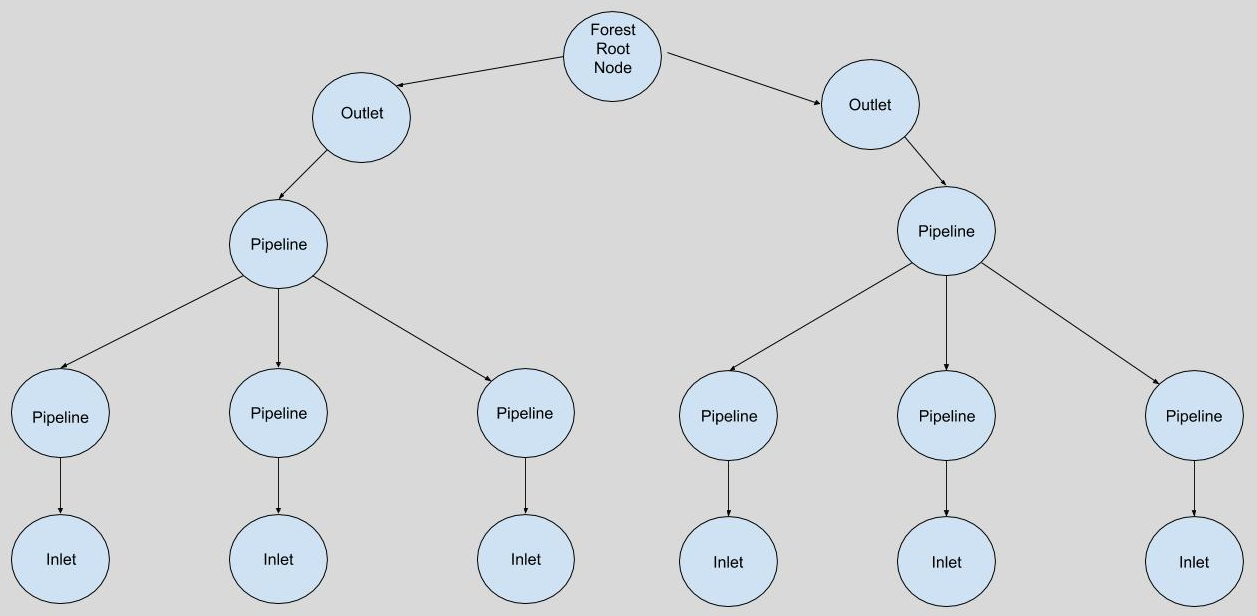}
\end{center}
   \caption{An example of tree-like data structure creation for capturing the process flow of pipeline schematics of P\&ID.}
\label{fig:tree_pid}
\end{figure} 

\subsection{Association and Structuring}
\label{subsec:association}
At this stage, we have detected all the necessary components of the P\&ID diagrams. Next step is to associate these components with each other and form a structure of the pipeline schematics. This is done as follows :
\begin{itemize}
\item \textit{Tags to Pipeline Association} : We find the line emerging direction from the orientation of inlet and outlet. We associate the closest pipeline from the line emerging point in the direction of pipeline to the tag. The closest pipeline is determined based upon euclidean distance.

\item \textit{Pipeline Code to Pipeline Association} : Similarly, we assign the pipeline codes to the nearest pipeline based on the minimum euclidean distance from any vertex of the bounding box of nearest  to the nearest point on the line.

\item \textit{Symbols to Pipeline Association} : Subsequently, every detected symbol will be associated to closest pipeline using minimum euclidean distance, provided it is not separated from the pipeline. 
\end{itemize}

Following this, we represent the structure of P\&ID diagrams in the form of a forest, as shown in Figure~\ref{fig:tree_pid}. Each outlet is treated as the root node of a specific tree in the forest and inlets are treated as leaf nodes. This means that all the lines are intermediate nodes. Each tree has minimum height of 2, root node has single child. Trees can have common nodes i.e., it can have common pipelines and inlet tags, but a root node is unique in the forest. At any time, a single flow path is represented by unique path between outlet and inlet.\\ 

\noindent \textbf{Tree Pruning} : The Pruning of tree is required to remove the false detections of pipelines by hough lines transform algorithm. A false positive pipeline is one which is represented in tree as a leaf node and does not link to any of the inlets. Therefore, we prune the tree by starting from the root node and removing all the nodes that do not lead to any inlet.

\begin{figure*}[t]
\begin{center}
   \includegraphics[width=0.9\linewidth]{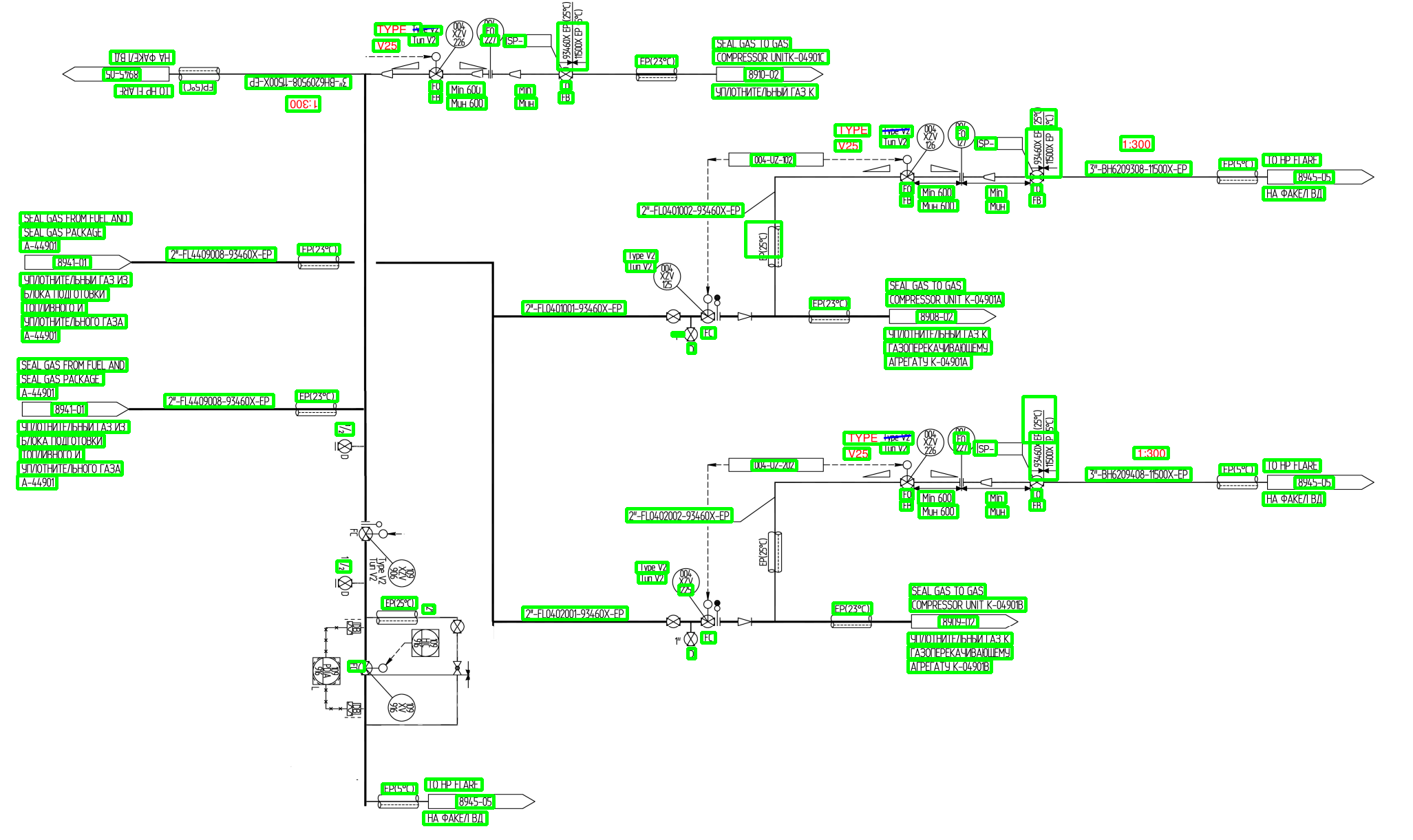}
\end{center}
   \caption{Figure showing text-detection output of pre-trained CTPN~\cite{ctpn}on P\&ID sheet.}
\label{fig:ctpn-res}
\end{figure*} 

\section{Experimental Results and Discussions}
\label{sec:experiment-results}
In this section, we evaluate the performance of our proposed end-to-end pipeline for extracting information from P\&ID sheets.
We use a dataset of real world PID sheets for quantitative evaluation which contains 4 number of sheets consisting of 672 flow diagrams. Table~\ref{tab:results-pipeline} shows the accuracy of detection and association of every component of the pipeline schematics. Row 1 of Table~\ref{tab:results-pipeline} gives the accuracy of pipeline code detection by CTPN followed by filtering of false positives using domain knowledge of standard code format. 64 codes are successfully detected out of total 71 giving accuracy of 90.1\%. We also show the visual output of CTPN on text detection on a sample P\&ID sheet, as given in Figure~\ref{fig:ctpn-res}. 

\begin{table*}[h]
\caption{Results of proposed pipeline for individual components. }
\centering{
\begin{tabular}{| l | c | c | c | c |}
\hline
& \multicolumn{2}{|c|}{\textbf{Results of}} \\
& \multicolumn{2}{|c|}{\textbf{individual component}}\\
\hline
\textbf{Component} & \textbf{Successful cases} & \textbf{Accuracy}\\
\hline
Pipeline-Code Detection &  $64$ / $71$  &  $90.1\%$ \\
\hline
Pipeline Detection &  $47$ / $72$ & $65.2\%$  \\
\hline
Outlet Detection &  $21$ / $21$ & $100\%$  \\
\hline
Inlet Detection &  $32$ / $32$ & $100\%$ \\
\hline
Pipeline Code Association &  $41$ / $64$ & $64.0\%$\\
\hline
Outlet Association &  $14$ / $21$ & $66.5\%$\\
\hline
Inlet Association &  $31$ / $32$ & $96.8\%$\\
\hline
\end{tabular}
}\\
\label{tab:results-pipeline}
\end{table*}

Next, pipelines are detected with an accuracy of 65.2\% because of some random noise such as line markings and overlaid diagrams. The proposed heuristics based method for outlet and inlet detection performed really well giving 100\% accuracy of detection, as given by Row 3 and 4, respectively. During the association of pipeline codes and outlets with the appropriate pipe, we were able to successfully associate 41 out of 64 pipeline codes and 14 out of 21 outlets, only. This is because of the fact that sometimes pipelines are not detected properly or pipelines do not intersect with the outlet, which happened in our case, as evident by pipeline detection accuracy given in Row 2 of Table~\ref{tab:results-pipeline}. However, inlets are associated quite successfully with the appropriate pipeline, giving an association accuracy of 96.8\%.

\begin{figure}[t]
\begin{center}
   \includegraphics[width=0.95\linewidth]{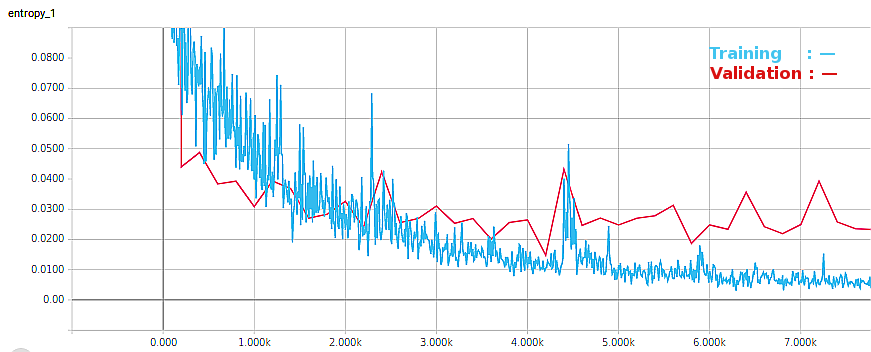}
\end{center}
   \caption{Plot showing cross-entropy loss for train and validation sets during training of FCN~\cite{shelhamerLD16} for symbol detection.}
\label{fig:fcn-training}
\end{figure}

\begin{table*}[h]
\caption{Confusion Matrix for Symbol Detection using FCN~\cite{shelhamerLD16} network.}
\label{tab:confusion-matrix}
\begin{center}
\small
\begin{tabular}{|c|c|c|c|c|c|c|c|c|c|c|c|}
\hline
& \multicolumn{11}{|c|}{\textit{\textbf{Predictions}}} \\
\hline
\textit{\textbf{Actual}}      & \textbf{Bl-V} & \textbf{Ck-V} & \textbf{Ch-sl} & \textbf{Cr-V} & \textbf{Con} & \textbf{F-Con} & \textbf{Gt-V-nc} & \textbf{Gb-V} & \textbf{Ins} & \textbf{Gb-V-nc} & \textbf{Others}\\
\hline
\textbf{Bl-V}   & \textbf{74}   &  2   &  0  & 0   & 0   & 0  & 0   &  4  &  0   &  0  &  0\\
\hline
\textbf{Ck-V}   &  0   &  \textbf{64}  &  0  & 0   & 4   & 0  & 0  &  0  &  0   &  0  &  0\\
\hline
\textbf{Ch-sl}   &  0   &  0   &  \textbf{25} & 0   & 0   & 0  & 0   &  0  &  0   &  0  &  0\\  
\hline
\textbf{Cr-V}   &  0   &  0   &  0  & \textbf{294} & 0   & 0  & 0   &  0  &  0   &  0  &  0\\
\hline
\textbf{Con}    &  0   &  0   &  0  & 0   & \textbf{38}  & 0  & 0   &  0  &  0   &  0  &  0\\
\hline
\textbf{F-Con}  &  0   &  0   &  0  & 0   & 0   & \textbf{41} & 0   &  0  &  0   &  1  &  0\\
\hline
\textbf{Gt-V-nc}&  0   &  0   &  0  & 0   & 0   & 8  & \textbf{36}  &  0  &  0   &  3  &  0\\
\hline
\textbf{Gb-V}   &  5   &  0   &  0  & 3   & 0   & 0  & 0   &  \textbf{64} &  0   &  0  &  0\\
\hline
\textbf{Ins}    &  0   &  0   &  0  & 0   & 0   & 0  & 0   &  0  &  \textbf{261} &  0  &  0\\
\hline
\textbf{Gb-V-nc}&  0   &  0   &  0  & 0   & 0   & 0  & 0   &  0  &  0   &  \textbf{52}  &  0\\
\hline
\textbf{Others} &  0   &  0   &  3  & 0   & 0   & 0  & 0   &  0  &  4   &  0  &  \textbf{149}\\
\hline
\end{tabular}
\end{center}
\end{table*}

Now, we present the results of symbol detection using FCN in the form of a confusion matrix, as shown in Table~\ref{tab:confusion-matrix}. FCN is trained for approx.7400 iterations and we saved the network at 7000 iterations by careful observation of the cross-entropy loss of train and validation set to prevent the network from overfitting. There are 10 different classes of symbols for detection in P\&ID sheets. We keep one extra class for training i.e. \textbf{Others} comprising of such symbols present in the P\&ID diagrams that are not of interest but were creating confusions in detection of symbols of interest. So, we have total of 11 classes of symbols for training FCN network for symbol detection.

\begin{figure}[h]
\begin{center}
   \includegraphics[width=0.95\linewidth]{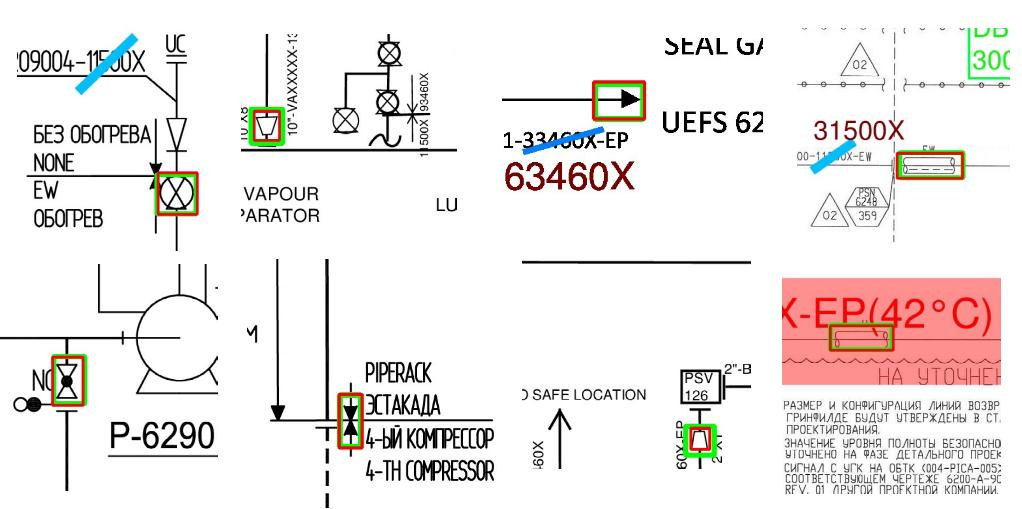}
\end{center}
   \caption{Examples of symbols detected using FCN. The green and red colored bounding boxes of symbols represent ground truth and corresponding predictions by FCN, respectively.}
\label{fig:fcn_output}
\end{figure}

\begin{table}[h]
\caption{Performance Measure of FCN~\cite{shelhamerLD16} on different classes of Symbol Detection.}
\label{tab:performance-fcn}
\begin{center}
\small
\begin{tabular}{|c|c|c|c|}
\hline
       & \textbf{Precision} & \textbf{Recall} & \textbf{F1-Score}\\
\hline
\textbf{Bl-V}   & 0.925   &  0.936 &  0.931 \\
\hline
\textbf{Ck-V}   &  0.941  &  0.969  &  0.955 \\
\hline
\textbf{Ch-sl}   &  1   &  0.893   &  0.944 \\  
\hline
\textbf{Cr-V}   &  1  &  0.989   &  0.995  \\
\hline
\textbf{Con}    &  1  &  0.905   &  0.95 \\
\hline
\textbf{F-Con}  &  0.976 &  0.837   &  0.901  \\
\hline
\textbf{Gt-V-nc}&  0.766 &  1   &  0.867  \\
\hline
\textbf{Gb-V}   &  0.888   &  0.941   &  0.914 \\
\hline
\textbf{Ins}    &  1   &  0.985   &  0.992 \\
\hline
\textbf{Gb-V-nc}&  1  &  0.929   &  0.963 \\
\hline
\textbf{Others} &  0.955   &  1   &  0.977  \\
\hline
\end{tabular}
\end{center}
\end{table}

We experimentally observe that FCN gives encouraging results for symbol detection with some minor confusions. As it is evident from the Figure~\ref{fig:symbols}, symbols such as ball\_valve, globe\_valve\_nc, gate\_valve\_nc, globe\_valve look visually similar and have very low inter-class variation in appearance. Most of the confusion is created among these classes of symbols only as given in Table~\ref{tab:confusion-matrix} with the exception of gate\_valve\_nc being recognised as flood\_connection which are not visually similar. For example, 5 out of 79 ball\_valve are being recognised as globe\_valve, 4 out of 68 globe\_valve are detected as ball\_valve, 3 out of 57 globe\_valve\_nc are recognized as gate\_valve\_nc. Symbols such as gate\_valve\_nc and concentric are detected successfully. We provide some sample examples of symbol detection using FCN in Figure~\ref{fig:fcn_output}.

We also calculate precision, recall and F1-score for each class of symbols, as given in Table~\ref{tab:performance-fcn}. We found that FCN detects symbols, even with very low visual difference in appearances, with impressive F1-scores of values more than 0.86 for every class. Precision is 100\% for symbols like chemical\_seal, circle\_valve, concentric, insulation and globe\_valve\_nc.

\section{Conclusion}
\label{sec:conclusion}
In this paper, we have proposed a novel end-to-end pipeline for extracting information from P\&ID sheets. We used state-of-the-art deep learning networks like CTPN and FCN for pipeline code and symbol detection, respectively and basic low level image processing techniques for detection of inlets, outlets and pipelines. We formulated a tree-like data structure for capturing the process flow of pipeline schematics after associating the detected components with the appropriate pipeline. We performed experiments on a dataset of real world P\&ID sheets using our proposed method and obtained satisfactory results. 

\bibliographystyle{apalike}
\bibliography{pid_icpram}

\vfill
\end{document}